\documentclass[runningheads]{llncs}

\usepackage[T1]{fontenc}
\def\doi#1{\href{https://doi.org/\detokenize{#1}}{\url{https://doi.org/\detokenize{#1}}}}
\usepackage{graphicx}
\usepackage{amsmath}

\usepackage{amsfonts}
\usepackage{todonotes}
\usepackage{listings}
\usepackage{multirow}
\usepackage{arydshln}
\usepackage{float}
\usepackage{textcomp}

\newcommand{\textapprox}{\raisebox{0.5ex}{\texttildelow}}

\begin{document}
\title{Encoding Surgical Videos as Latent Spatiotemporal Graphs for Object and Anatomy-Driven Reasoning}
\author{Aditya Murali\inst{1} \and
Deepak Alapatt\inst{1} \and
Pietro Mascagni\inst{2, 3} \and
Armine Vardazaryan\inst{2} \and
Alain Garcia\inst{2} \and
Nariaki Okamoto\inst{4} \and
Didier Mutter\inst{2} \and
Nicolas Padoy\inst{1, 2}}
%

\authorrunning{A. Murali et al.}
%
\institute{ICube, University of Strasbourg, CNRS, Strasbourg, France \and
IHU-Strasbourg, Institute of Image-Guided Surgery, Strasbourg, France \and
Fondazione Policlinico Universitario Agostino Gemelli IRCCS, Rome, Italy \and
Institute for Research Against Digestive Cancer (IRCAD), Strasbourg, France}
\titlerunning{Encoding Surgical Videos as Spatiotemporal Graphs}

\maketitle

\begin{abstract}
Recently, spatiotemporal graphs have emerged as a concise and elegant manner of representing video clips in an object-centric fashion, and have shown to be useful for downstream tasks such as action recognition.
In this work, we investigate the use of latent spatiotemporal graphs to represent a surgical video in terms of the constituent anatomical structures and tools and their evolving properties over time.
To build the graphs, we first predict frame-wise graphs using a pre-trained model, then add temporal edges between nodes based on spatial coherence and visual and semantic similarity.
Unlike previous approaches, we incorporate long-term temporal edges in our graphs to better model the evolution of the surgical scene and increase robustness to temporary occlusions.
We also introduce a novel graph-editing module that incorporates prior knowledge and temporal coherence to correct errors in the graph, enabling improved downstream task performance.
Using our graph representations, we evaluate two downstream tasks, critical view of safety prediction and surgical phase recognition, obtaining strong results that demonstrate the quality and flexibility of the learned representations. Code is available at github.com/CAMMA-public/SurgLatentGraph.

\end{abstract}

\section{Introduction}

Surgical video analysis is a rapidly growing field that aims to improve and gain insights into surgical practice by leveraging increasingly available surgical video footage~\cite{MaierHein2017,vercauteren2019cai4cai}.
Several key applications have been well explored, ranging from surgical skill assessment to workflow analysis to intraoperative safety enhancement~\cite{wu2021cross,FunkeBOBWS19,gao2021trans,madani2022artificial}.
Yet, effectively learning and reasoning based on surgical anatomy remains a challenging problem, as evidenced by lagging performance in fine-grained tasks such as surgical action triplet detection and critical view of safety prediction~\cite{nwoye2022cholectriplet2021,murali2022latent}.

\begin{figure}
\includegraphics[width=\textwidth]{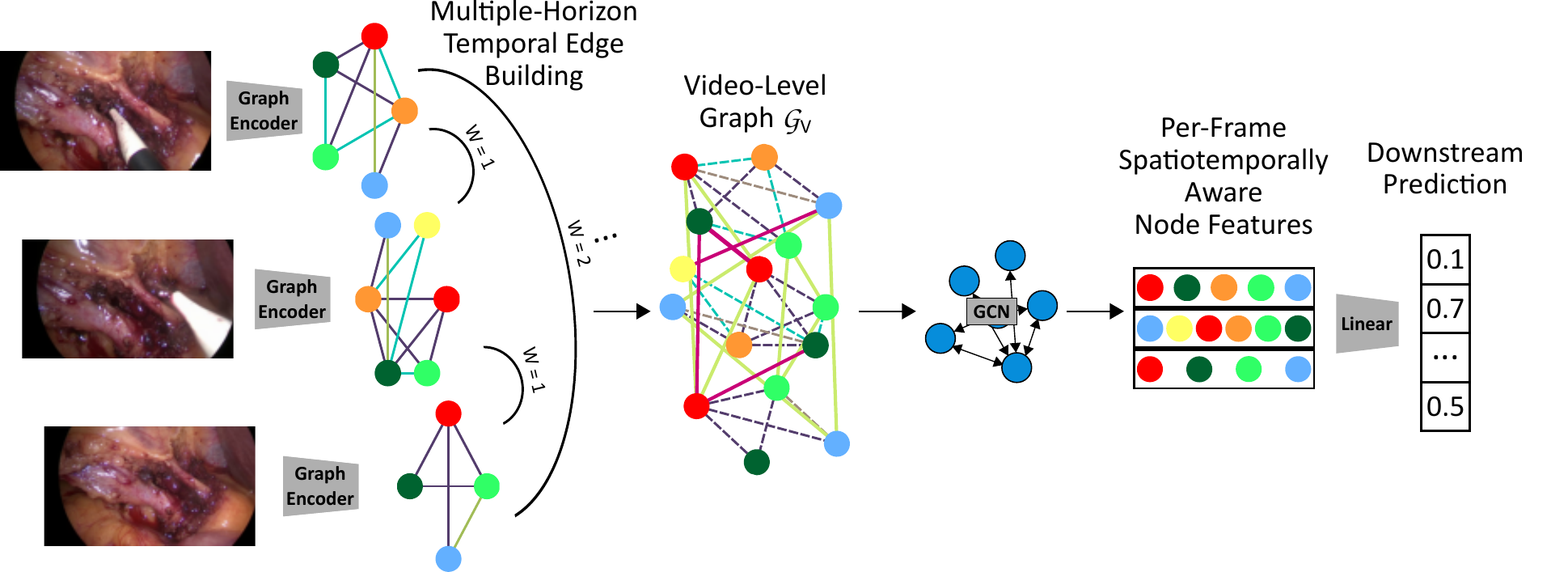}
\caption{Overview of our proposed approach. We begin by computing graphs for each frame using~\cite{murali2022latent}, then add temporal edges (shown w.ith solid lines) between graphs at different horizons to obtain the video-level graph $\mathcal{G}_V$. We process $\mathcal{G}_V$ with a GNN to yield spatiotemporally-aware node features, which we use for downstream prediction. Each node color corresponds to an object class, and each edge color to a relation class. We retain spatial edges (shown with dotted lines) from the graph encoder in $\mathcal{G}_V$.}
\label{fig:arch}
\end{figure}

Such anatomy-based reasoning can be accomplished through \textit{object-centric modeling}, which is gaining popularity in general computer vision~\cite{raboh2020differentiable,materzynska2020something,wang2018videos,Zhang_2022_ACCV,Herzig_2022_CVPR}.
Object-centric models represent images or clips according to their constituent objects by running an object detector and then using the detections to factorize the visual feature space into per-object features.
By retaining implicit visual features, these approaches maintain differentiability, allowing them to be fine-tuned for various downstream tasks.
Meanwhile, they can also be extended to include object attributes such as class, location, and temporal order for tasks that rely heavily on scene semantics.
Recent works have explored object-centric representations in the surgical domain, but they are characterized by one of several limitations: they often include only surgical tools~\cite{khan2021spatiotemporal,sarikaya2020towards}, preventing anatomy-driven reasoning; they are limited to single-frames or short clips~\cite{pang2022rethinking,murali2022latent,khan2021spatiotemporal,sarikaya2020towards}, preventing video-level understanding; or they formulate the object-centric representation as a final output (e.g. scene graph prediction) and only include scene semantics, which limits their effectiveness for downstream tasks~\cite{islam2020learning,seenivasan2022global,ozsoy20224d}.

In this work, we tackle these challenges by proposing to build latent spatiotemporal graph representations of \textit{entire surgical videos}, with each node representing a surgical tool or anatomical structure and edges representing relationships between nodes across space and time.
To build our graphs, rather than use an off-the-shelf object detector, we employ the latent graph encoder of~\cite{murali2022latent} to generate per-frame graphs that additionally encode object semantics, segmentation details, and inter-object relations, all of which are important for downstream anatomy-driven reasoning.
We then add edges between nodes in different graphs, resulting in a spatiotemporal graph representation of the entire video.
We encounter two main challenges when building these graphs for surgical videos: (1) surgical scenes evolve slowly over time, calling for long-term modeling, and (2) object detection is often error-prone due to annotated data scarcity. 
To address the first challenge, we introduce a framework to add temporal edges at multiple horizons, enabling reasoning about the short-term and long-term evolution of the underlying video.
Then, to address the error-prone object detection, we propose a Graph Editing Module that leverages the spatiotemporal graph structure and predicted object semantics to efficiently correct errors in object detection.

We evaluate our method on two downstream tasks: critical view of safety (CVS) clip classification and surgical phase recognition.
CVS clip classification is a fine-grained task that requires accurate identification and reasoning about anatomy, and is thus an ideal target application for our object-centric approach.
On the other hand, phase recognition is a coarse-grained task that requires holistic understanding of longer video segments, which can demonstrate the effectiveness of our temporal edge building framework.
We achieve competitive performance in both of these tasks and show that our graph representations can be used with or without task-specific finetuning, thereby demonstrating their value as general-purpose video representations.

In summary, we contribute the following:
\begin{enumerate}
    \item A method to encode surgical videos as latent spatiotemporal graphs that can then be used without modification for two diverse downstream tasks.
    \item A framework for effectively modeling long-range relationships in surgical videos via multiple-horizon temporal edges.
    \item A Graph Editing Module that can correct errors in the predicted graph based on temporal coherence cues and prior knowledge.
\end{enumerate}

\section{Methods}
\label{sec:methods}

In this section, we describe our approach to encode a $T$-frame video $V = \{I_t\ |\ 1 \leq t \leq T\}$ as a latent spatiotemporal graph $G_V$ (illustrated in Fig. \ref{fig:arch}).
Our method consists of a frame-wise object detection step followed by a temporal graph building step and a graph editing module to correct errors in the predicted graph representation.
We also describe our graph neural network decoder to process the resulting representation $G_V$ for downstream tasks.

\subsection{Graph Construction}
\label{subsec:st_graph_construction}

%
\textbf{Object Detection.} To construct a latent spatiotemporal graph representation, we must first detect the objects in each frame, along with any additional properties.
We do so by employing the graph encoder $\phi_{\text{SG}}$ proposed in~\cite{murali2022latent}, yielding a graph $\mathcal{G}_t$ for each frame $I_t \in V$.
The resulting $\mathcal{G}_t$ is composed of nodes $\mathcal{N}_t$ and edges $\mathcal{E}_t$; $\mathcal{N}_t$ and $\mathcal{E}_t$ are in turn composed of features $h_i$, $h_{i, j}$, bounding boxes $b_i$, $b_{i, j}$, and object/relation class $r_i$, $r_{i, j}$.

\noindent \textbf{Spatiotemporal Graph Building.} Once we have computed the graphs $\mathcal{G}_t$, we add temporal edges to construct a single graph $\mathcal{G}_V$ that describes the entire video.
$\mathcal{G}_V$ retains the spatial edges from the various $\mathcal{G}_t$ to describe geometric relations between objects in the same frame (i.e. to the left of, above), while also including temporal edges between spatially and visually similar nodes.
It can then be processed with a graph neural network during downstream evaluation to efficiently propagate object-level information across space and time.

We add temporal edges to $\mathcal{G}_V$ based on object bounding box overlap and visual feature similarity, inspired by~\cite{wang2018videos}; however, we extend their approach to construct edges at multiple temporal horizons rather than between adjacent frames alone.
Specifically, we design an operator $\phi_{\text{TE}}$ that takes a pair of graphs $G_t, G_{t+w}$ and outputs a list of edges, which are defined by their connectivity $\mathcal{C}_{t, t+w}$ containing pairs of adjacent nodes, and their relation class $\mathcal{R}_{t, t+w}$ containing relation class ids.
To compute the edges, $\phi_{\text{TE}}$ calculates pairwise similarities between nodes in $G_t$ and nodes in $G_{t+w}$ using two separate kernels: $K_B$, which computes the generalized IoU between node bounding boxes, and $K_F$, which computes the cosine similarity between node features.
This yields similarity matrices $M_B$ and $M_F$, each of size $\mathcal{N}_t \times \mathcal{N}_{t+w}$.
Using each matrix, we select the most similar node $n_{j, t+w} \in \mathcal{N}_{t+w}$ for each $n_{i, t} \in \mathcal{N}_t$ and vice-versa.
Altogether, this yields $4 * \left(|\mathcal{N}_t| + |\mathcal{N}_{t+w}|\right)$ edges consisting of connectivity tuples $c_{m, n} = \left((i, t), (j, t+w)\right)$ and relation classes $r_{m, n}$, which we store in $\mathcal{C}_{t, t+w}$ and $\mathcal{R}_{t, t+w}$ respectively. 
We apply $\phi_{\text{TE}}$ to all pairs of graphs $\mathcal{G}_t, \mathcal{G}_{t+w}$ for various temporal horizons $w \in \mathcal{W}$, then combine the resulting $\mathcal{C}_{t, t+w}$ and $\mathcal{R}_{t, t+w}$ to obtain temporal edge connectivities $\mathcal{C}_\text{ST}$ and relation classes $\mathcal{R}_\text{ST}$.
Finally, we augment each temporal edge with features $h_{m, n}$ and bounding boxes $b_{m, n}$, yielding the video-level graph $\mathcal{G}_V$:
\begin{equation}
\centering
\begin{gathered}
    h_{m, n} = h_{i, t_x} + h_{j, t_y},\ b_{m, n} = \cup(b_{i, t_x}, b_{j, t_y}),\ \text{where}\ (i, t_x), (j, t_y) = c_{m, n}\\
    \mathcal{E}_{\text{ST}} = \{b_{m, n}, r_{m, n}, h_{m, n}\ |\ (m, n) \in \mathcal{C}_{\text{ST}}\};\ \mathcal{E}_V = \{\mathcal{E}_t\ |\ 1 \leq t \leq T\} \cup \mathcal{E}_{\text{ST}}\\
    \mathcal{G}_{V} = \{\mathcal{N}_V, \mathcal{E}_V\},\ \text{where}\ \mathcal{N}_V = \bigcup_{1 \leq t \leq T} \mathcal{N}_t.
\end{gathered}
\end{equation}

\noindent \textbf{Edge Horizon Selection.} While $\phi_{\text{TE}}$ is designed to construct edges between arbitrarily distant graphs, effective selection of temporal horizons $\mathcal{W}$ is non-trivial.
We could naively include every possible temporal horizon, setting $\mathcal{W} = \{1, 2, ..., T-1\}$ to maximize temporal information flow; however, making $\mathcal{W}$ too dense results in redundancies in the resulting graph, which can have an oversmoothing effect during downstream processing with a graph neural network (GNN)~\cite{zhang2019graph}.
To avoid this issue, we take inspiration from temporal convolutional networks (TCN)~\cite{lea2017temporal}, which propagate information over long input sequences using a series of convolutions with exponentially increasing dilation.
We similarly use exponentially increasing temporal horizons, setting $\mathcal{W} = \{1, 2, 4, ..., 2^l\}$ to enable efficient information flow at each GNN layer and long-horizon message passing via a stack of GNN layers.

\subsection{Graph Editing Module}
\label{subsec:st_graph_construction}
One limitation of object-centric approaches is a reliance on high quality object detection, which is particularly steep in surgical videos.
These difficulties in object detection could be tackled by incorporating prior knowledge such as anatomical scene geometry, but incorporating these constraints into the learning process often requires complex constraint formulations and methodologies.
We posit that our spatiotemporal graph structure represents a simpler framework to incorporate such constraints; to demonstrate this, we introduce a module to filter detections of anatomical structures, which are particularly difficult to detect, incorporating the constraint that there is only one of each structure in each frame.
Specifically, after building the spatiotemporal graph, we compute a dropout probability $p_{i, t} = \frac{1}{deg(n_{i, t})}$ for each node, where $deg$ is the degree operator.
Then, for each frame $t$, for each object class $r_j$, we select the highest scoring node $n_t$ from $\{n_{i, t} | r_{i, t}\}$.
During training, we apply graph editing with probability $p_{\text{edit}}$, providing robustness to a wide range of input graphs.

\subsection{Downstream Task Decoder}
\label{subsec:downstream_task_decoder}
For downstream prediction from $\mathcal{G}_V$, we first apply a GNN using the architecture proposed in~\cite{dhamo2020semantic}, yielding spatiotemporally-aware node features.
Then, we pool the node features within each frame and apply a linear layer to yield frame-wise predictions (see Fig. \ref{fig:arch}).
We process these predictions differently depending on the task: for clip classification, we output only the prediction for the last frame, while for temporal video segmentation, we output the frame-wise predictions.

\subsection{Training}
\label{subsec:training}
We adopt a two-stage training approach, starting by training $\phi_{\text{SG}}$ as proposed in~\cite{murali2022latent} and then extracting graphs for all images.
Then, in the second stage, we process a sequence of graphs with our model to predict frame-wise outputs.
We supervise each prediction with the corresponding frame label, propagating the clip label to each frame when per-frame labels are unavailable.

\section{Experiments and Results}
\label{sec:experiments}
In this section, we describe our evaluation tasks and datasets, describe baseline methods and our model, then present results for each task.
We conclude with an ablation study that illustrates the impact of our various model components.

\subsection{Evaluation Tasks and Datasets}
\label{subsec:ds_tasks_and_datasets}

\textbf{Critical View of Safety (CVS) Prediction.} The CVS consists of three independent criteria, and can be viewed as a multi-label classification problem~\cite{mascagni2021artificial}.
For our experiments, we use the Endoscapes+ dataset introduced in~\cite{murali2022latent}, which contains 11090 images annotated with CVS evenly sampled from the dissection phase of 201 cholecystectomies at 0.2 fps; it also includes a subset of 1933 images with segmentation masks and bounding box annotations.
We model CVS prediction as a clip classification problem, constructing clips of length 10 at 1 fps, and use the label of the last frame as the clip label.
As in~\cite{murali2022latent}, we investigate CVS prediction performance in two experimental settings to study the label-efficiency of various methods: (1) using only the bounding box labels and CVS labels and (2) additionally using the segmentation labels.
We report mean average precision (mAP) across the three criteria for all methods.

\noindent \textbf{Surgical Phase Recognition.} For surgical phase recognition, we use the publically available Cholec80 dataset~\cite{twinanda2016endonet}, which includes 80 videos with frame-wise phase annotations ($\{1,2,...,7\}$).
We use the first 40 videos for training, the next 8 for validation, and the remaining 32 for testing, as in~\cite{czempiel2020tecno,ramesh2022dissecting}.
In addition, to enable object-centric approaches, we use the publically available CholecSeg8k dataset~\cite{hong2020cholecseg8k} as it represents multiple surgical phases unlike Endoscapes+.
As CholecSeg8k is a subset of Cholec80, we split the images into training, validation, and testing following the aforementioned video splits.
We model phase recognition as a temporal video segmentation problem, and process the entire video at once.
Again, we explore two experimental settings: (1) temporal phase recognition without single-frame finetuning to evaluate the surgical video representations learned by each method and (2) temporal phase recognition with single-frame finetuning, the classical setting~\cite{czempiel2020tecno,gao2021trans,ramesh2022dissecting}.
We report mean F1 score across videos for all methods, as suggested in~\cite{ramesh2022dissecting}.

\subsection{Baseline Methods}
\noindent\textbf{Single-Frame Methods.} As CVS clip classification is unexplored, we compare to two recent single-frame methods for reference: LG-CVS~\cite{murali2022latent}, a graph-based approach, and DeepCVS~\cite{mascagni2021artificial}, a non-object-centric approach. We quote results from~\cite{murali2022latent}, which improves DeepCVS and enables training with bounding boxes.

\noindent \textbf{DeepCVS-Temporal.} We also extend DeepCVS for clip classification by replacing the last linear layer of the dilated ResNet-18 with a Transformer decoder, referring to this model as DeepCVS-Temporal.

\noindent \textbf{STRG}. Space-Time Region Graphs (STRG)~\cite{wang2018videos} is a spatiotemporal graph-based approach for action recognition that builds a latent graph by predicting region proposals and extracting per-region visual features using an I3D backbone; we repurpose STRG for CVS clip classification and phase recognition.
Because STRG is trained end-to-end, it can only process clips of limited length; consequently, we train STRG on clips of $15$ frames for phase recognition rather the entire video as in other methods.
We also only consider phase recognition with finetuning.
For CVS clip classification, we additionally pre-train the I3D feature extractor in STRG on bounding box/segmentation labels using a FasterRCNN box head~\cite{ren2015faster} or DeepLabV3+ decoder~\cite{Chen_2018_ECCV}.

\noindent \textbf{TeCNO.} TeCNO~\cite{czempiel2020tecno} is a temporal model for surgical phase recognition consisting of frame-wise feature extraction followed by temporal decoding with a causal TCN~\cite{lea2017temporal} to classify phases.
For phase recognition without single-frame finetuning, we use a ResNet-50 pre-trained on CholecSeg8k using a DeepLabV3+ head to extract features, enabling fair comparisons with our method.
For the other setting, we report performance from~\cite{ramesh2022dissecting}.

\noindent \textbf{Ours.} We train our model in two stages, starting by training the graph encoder $\phi_{\text{SG}}$ as described in~\cite{murali2022latent} on the subset of Endoscapes+ annotated with segmentation masks or bounding boxes for CVS clip classification, or on CholecSeg8k for phase recognition.
We then extract frame-wise graphs for the entire dataset and apply our spatiotemporal graph approach to predict CVS or phase.
In the second experimental setting for phase recognition, we additionally finetune $\phi_{\text{SG}}$ on all training images with the frame-wise phase labels before extracting the graphs. 
Finally, we evaluate a version of our method that additionally applies a TCN to the un-factorized image features and adds the TCN-processed features to the pooled temporally-aware node features prior to linear classification.
We set $l = 3$, $p_{\text{edit}} = 0.5$, and use a 5-layer GNN for CVS prediction and an 8-layer GNN for phase recognition.

\subsection{Main Experiments}
\label{subsec:main_exps}

\begin{table}[t]
\centering
\caption{CVS Clip Classification Performance. Standard deviation is across three runs of each method. Single frame methods from prior works are also reported for reference.}
\label{tab:cvs_prediction}
\renewcommand{\arraystretch}{1}
\begin{tabular*}{\linewidth}{@{\extracolsep{\fill}} cccc }
\multicolumn{2}{c}{}                                     & \multicolumn{2}{c}{CVS mAP}                          \\
\multicolumn{2}{c}{\multirow{-2}{*}{Method}}             & Box                                  & Seg           \\ \hline
                               & DeepCVS-R18~\cite{mascagni2021artificial,murali2022latent} & $54.1 \pm 1.3$ & $60.2 \pm 1.6$ \\
\multirow{-2}{*}{Single Frame} & LG-CVS~\cite{murali2022latent} & $63.6 \pm 0.8$                        & $67.3 \pm 1.4$ \\ \hline
                               & DeepCVS-Temporal & $57.8 \pm 3.2$                                & $63.8 \pm 2.1$ \\
                               & STRG~\cite{wang2018videos}  & $60.5 \pm 0.7$ & $61.7 \pm 1.5$ \\
\multirow{-3}{*}{Temporal}     & Ours                    & $\mathbf{66.3 \pm 0.9}$                        & $\mathbf{69.7 \pm 1.3}$
\end{tabular*}
\end{table}

\textbf{CVS.}
Our first takeaway from Table \ref{tab:cvs_prediction} is that temporal models provide a method-agnostic boost of \textapprox$3\%$ mAP for CVS prediction.
Furthermore, our approach outperforms both non-object-centric and object-centric temporal baselines, achieving a substantial performance boost in the label-efficient bounding box setting while remaining competitive when also trained with segmentation masks.
In the box setting, we observe that the non-object-centric DeepCVS approaches perform rather poorly due to an over-reliance on predicted semantics rather than effective visual encodings~\cite{murali2022latent}.
Object-centric modeling addresses some of these limitations, as evidenced by STRG outperforming DeepCVS-Temporal.
Nevertheless, our method achieves a much stronger performance boost, owing to multiple factors: (1) our model is based on the underlying LG-CVS, which constructs its frame-wise object-centric representation by using the final bounding box predictions rather than just region proposals like STRG, and also encodes semantic information, and (2) our proposed improvements (multiple-horizon edges, graph editing) are critical to improving model performance.
Meanwhile, in the segmentation setting, the object-centric STRG is ineffective, performing worse than DeepCVS-Temporal; this discrepancy arises because, as previously mentioned, STRG relies on region proposals rather than object-specific bounding boxes in its graph representation, and as a result, cannot fully take advantage of the additional information provided by the segmentation masks. 
Our approach translates the ideas of STRG but importantly builds on top of the already effective representations learnt by LG-CVS to achieve universal effectiveness for spatiotemporal modeling of CVS.

\begin{table}[t]
\centering
\caption{Surgical Phase Recognition Performance.}
\label{tab:phase_recognition}
\renewcommand{\arraystretch}{1}
\begin{tabular*}{\linewidth}{@{\extracolsep{\fill}} ccc }
\multicolumn{2}{c}{Method} & Phase F1 \\ \hline
 & TeCNO~\cite{czempiel2020tecno} & $64.3$ \\
 & Ours & $70.3$ \\ 
\multirow{-3}{*}{\begin{tabular}[c]{@{}c@{}}No Single-Frame Finetuning\end{tabular}} & Ours + TCN & $\mathbf{74.1}$ \\ \hline
 & STRG~\cite{wang2018videos} & $77.1$ \\
 & TeCNO~\cite{czempiel2020tecno}, reported from~\cite{ramesh2022dissecting} & $80.3$ \\
 & Ours & $79.9$ \\
\multirow{-4}{*}{\begin{tabular}[c]{@{}c@{}}With Single-Frame Finetuning\end{tabular}} & Ours + TCN & $\mathbf{81.4}$
\end{tabular*}
\end{table}

\noindent \textbf{Phase.} Table \ref{tab:phase_recognition} shows the phase recognition results for various methods with (bottom) and without (top) finetuning the underlying single-frame model.
Our model is already highly effective for phase recognition without any finetuning, outperforming the corresponding TeCNO model by $6.1$\% F1 in its original form and by nearly $10\%$ F1 when also using a TCN.
This shows that the graph representations contain general-purpose information about the surgical scenes and their evolution.
Finally, by finetuning the underlying single-frame graph encoder, we match the existing state-of-the-art, highlighting our method's flexibility.

\begin{table}[t]
\centering
\caption{Ablation Study of Model Components.}
\label{tab:ablation}
\begin{tabular*}{\linewidth}{@{\extracolsep{\fill}} ccc }
\multirow{2}{*}{Ablated Feature} &\multicolumn{2}{c}{Performance Drop ($\uparrow$ is worse)} \\
 & CVS mAP (Seg) & Phase F1 (No FT) \\ \hline
No Long Term Edges ($\mathcal{W} = \{1\}$) & 1.6 & 7.2 \\
\begin{tabular}[c]{@{}c@{}}Naive Edge Horizon Selection\\ $\left(\mathcal{W} = \{1,2,...,T-1\}\right)$ \end{tabular}
& 1.4 & 4.1 \\
No Graph Editing Module & 1.1 & 0.2
\end{tabular*}
\end{table}

\subsection{Ablation Studies}
\label{subsec:ablations}
\noindent Table \ref{tab:ablation} illustrates the impact of each model component on CVS clip classification and phase recognition performance.
The first two rows illustrate the importance of using exponential edge horizons.
Without any long-term edges (as in STRG), we observe a staggering $7.20\%$ drop in Phase F1; naively building edges between all the graphs improves performance but is still $4.14\%$ worse than our proposed method.
We observe similar trends for the CVS mAP but with lower magnitude, as CVS prediction is not as reliant on long-term video understanding.
Meanwhile, we observe the opposite effect for the graph editing module, which is quite effective for CVS clip classification but does not considerably impact phase F1.
This is again consistent with the nature of the tasks, as CVS requires fine-grained understanding of the surgical anatomy, and performance can suffer greatly from errors in object detection, while phase recognition is more coarse-grained and is thus less impacted by errors at this fine-grained level.

\section{Conclusion}
\label{sec:conclusion}
We introduce a method to encode surgical videos in their entirety as latent spatiotemporal graph representations.
Our graph representations enable fine-grained anatomy-driven reasoning as well as coarse long-range video understanding due to the inclusion of edges at multiple-temporal horizons, robustness against errors in object detection provided by a graph editing module, and memory- and computational-efficiency afforded by a two-stage training pipeline.
We believe that the resulting graphs are powerful general-purpose representations of surgical videos that can fuel numerous future downstream applications.

\section{Acknowledgement}
This work was supported by French state funds managed by the ANR within the National AI Chair program under Grant ANR-20-CHIA-0029-01 (Chair AI4ORSafety) and within the Investments for the future program under Grants ANR-10-IDEX-0002-02 (IdEx Unistra) and ANR-10-IAHU-02 (IHU Strasbourg). This work was granted access to the HPC resources of IDRIS under the allocation 2021-AD011011640R1 made by GENCI.

\bibliographystyle{splncs04}
\bibliography{main}

\begin{thebibliography}{10}
\providecommand{\url}[1]{\texttt{#1}}
\providecommand{\urlprefix}{URL }
\providecommand{\doi}[1]{https://doi.org/#1}

\bibitem{Chen_2018_ECCV}
Chen, L.C., Zhu, Y., Papandreou, G., Schroff, F., Adam, H.: Encoder-decoder
  with atrous separable convolution for semantic image segmentation. In:
  Proceedings of the European Conference on Computer Vision (ECCV) (September
  2018)

\bibitem{czempiel2020tecno}
Czempiel, T., Paschali, M., Keicher, M., Simson, W., Feussner, H., Kim, S.T.,
  Navab, N.: Tecno: Surgical phase recognition with multi-stage temporal
  convolutional networks. In: Medical Image Computing and Computer Assisted
  Intervention. pp. 343--352. Springer (2020)

\bibitem{dhamo2020semantic}
Dhamo, H., Farshad, A., Laina, I., Navab, N., Hager, G.D., Tombari, F.,
  Rupprecht, C.: Semantic image manipulation using scene graphs. In: CVPR. pp.
  5213--5222 (2020)

\bibitem{FunkeBOBWS19}
Funke, I., Bodenstedt, S., Oehme, F., von Bechtolsheim, F., Weitz, J., Speidel,
  S.: Using {3D} convolutional neural networks to learn spatiotemporal features
  for automatic surgical gesture recognition in video. In: Medical Image
  Computing and Computer Assisted Intervention (2019)

\bibitem{gao2021trans}
Gao, X., Jin, Y., Long, Y., Dou, Q., Heng, P.A.: Trans-svnet: Accurate phase
  recognition from surgical videos via hybrid embedding aggregation
  transformer. In: Medical Image Computing and Computer Assisted Intervention.
  pp. 593--603. Springer (2021)

\bibitem{Herzig_2022_CVPR}
Herzig, R., Ben-Avraham, E., Mangalam, K., Bar, A., Chechik, G., Rohrbach, A.,
  Darrell, T., Globerson, A.: Object-region video transformers. In: Proceedings
  of the IEEE/CVF Conference on Computer Vision and Pattern Recognition (CVPR).
  pp. 3148--3159 (June 2022)

\bibitem{hong2020cholecseg8k}
Hong, W.Y., Kao, C.L., Kuo, Y.H., Wang, J.R., Chang, W.L., Shih, C.S.:
  Cholecseg8k: a semantic segmentation dataset for laparoscopic cholecystectomy
  based on cholec80. arXiv preprint arXiv:2012.12453  (2020)

\bibitem{islam2020learning}
Islam, M., Seenivasan, L., Ming, L.C., Ren, H.: Learning and reasoning with the
  graph structure representation in robotic surgery. In: Medical Image
  Computing and Computer Assisted Intervention. pp. 627--636. Springer (2020)

\bibitem{khan2021spatiotemporal}
Khan, S., Cuzzolin, F.: Spatiotemporal deformable scene graphs for complex
  activity detection. BMVC  (2021)

\bibitem{lea2017temporal}
Lea, C., Flynn, M.D., Vidal, R., Reiter, A., Hager, G.D.: Temporal
  convolutional networks for action segmentation and detection. In: proceedings
  of the IEEE Conference on Computer Vision and Pattern Recognition. pp.
  156--165 (2017)

\bibitem{madani2022artificial}
Madani, A., Namazi, B., Altieri, M.S., Hashimoto, D.A., Rivera, A.M., Pucher,
  P.H., Navarrete-Welton, A., Sankaranarayanan, G., Brunt, L.M., Okrainec, A.,
  et~al.: Artificial intelligence for intraoperative guidance: using semantic
  segmentation to identify surgical anatomy during laparoscopic
  cholecystectomy. Annals of surgery  (2022)

\bibitem{MaierHein2017}
Maier-Hein, L., Vedula, S.S., Speidel, S., Navab, N., Kikinis, R., Park, A.,
  Eisenmann, M., Feussner, H., Forestier, G., Giannarou, S., Hashizume, M.,
  Katic, D., Kenngott, H., Kranzfelder, M., Malpani, A., Marz, K., Neumuth, T.,
  Padoy, N., Pugh, C., Schoch, N., Stoyanov, D., Taylor, R., Wagner, M., Hager,
  G.D., Jannin, P.: Surgical data science for next-generation interventions.
  Nature Biomedical Engineering  \textbf{1}(9),  691--696 (Sep 2017)

\bibitem{mascagni2021artificial}
Mascagni, P., Vardazaryan, A., Alapatt, D., Urade, T., Emre, T., Fiorillo, C.,
  Pessaux, P., Mutter, D., Marescaux, J., Costamagna, G., et~al.: Artificial
  intelligence for surgical safety: automatic assessment of the critical view
  of safety in laparoscopic cholecystectomy using deep learning. Annals of
  Surgery  (2021)

\bibitem{materzynska2020something}
Materzynska, J., Xiao, T., Herzig, R., Xu, H., Wang, X., Darrell, T.:
  Something-else: Compositional action recognition with spatial-temporal
  interaction networks. In: CVPR. pp. 1049--1059 (2020)

\bibitem{murali2022latent}
Murali, A., Alapatt, D., Mascagni, P., Vardazaryan, A., Garcia, A., Okamoto,
  N., Mutter, D., Padoy, N.: Latent graph representations for critical view of
  safety assessment. arXiv preprint arXiv:2212.04155  (2022)

\bibitem{nwoye2022cholectriplet2021}
Nwoye, C.I., Alapatt, D., Yu, T., Vardazaryan, A., Xia, F., Zhao, Z., Xia, T.,
  Jia, F., Yang, Y., Wang, H., et~al.: Cholectriplet2021: A benchmark challenge
  for surgical action triplet recognition. arXiv preprint arXiv:2204.04746
  (2022)

\bibitem{ozsoy20224d}
{\"O}zsoy, E., {\"O}rnek, E.P., Eck, U., Czempiel, T., Tombari, F., Navab, N.:
  4d-or: Semantic scene graphs for or domain modeling. arXiv preprint
  arXiv:2203.11937  (2022)

\bibitem{pang2022rethinking}
Pang, W., Islam, M., Mitheran, S., Seenivasan, L., Xu, M., Ren, H.: Rethinking
  feature extraction: Gradient-based localized feature extraction for
  end-to-end surgical downstream tasks. IEEE Robotics and Automation Letters
  \textbf{7}(4),  12623--12630 (2022)

\bibitem{raboh2020differentiable}
Raboh, M., Herzig, R., Berant, J., Chechik, G., Globerson, A.: Differentiable
  scene graphs. In: Proceedings of the IEEE/CVF Winter Conference on
  Applications of Computer Vision. pp. 1488--1497 (2020)

\bibitem{ramesh2022dissecting}
Ramesh, S., Srivastav, V., Alapatt, D., Yu, T., Murali, A., Sestini, L., Nwoye,
  C.I., Hamoud, I., Fleurentin, A., Exarchakis, G., et~al.: Dissecting
  self-supervised learning methods for surgical computer vision. arXiv preprint
  arXiv:2207.00449  (2022)

\bibitem{ren2015faster}
Ren, S., He, K., Girshick, R., Sun, J.: Faster r-cnn: Towards real-time object
  detection with region proposal networks. Advances in neural information
  processing systems  \textbf{28} (2015)

\bibitem{sarikaya2020towards}
Sarikaya, D., Jannin, P.: Towards generalizable surgical activity recognition
  using spatial temporal graph convolutional networks. arXiv preprint
  arXiv:2001.03728  (2020)

\bibitem{seenivasan2022global}
Seenivasan, L., Mitheran, S., Islam, M., Ren, H.: Global-reasoned multi-task
  learning model for surgical scene understanding. IEEE Robotics and Automation
  Letters  \textbf{7}(2),  3858--3865 (2022)

\bibitem{twinanda2016endonet}
Twinanda, A.P., Shehata, S., Mutter, D., Marescaux, J., De~Mathelin, M., Padoy,
  N.: Endonet: a deep architecture for recognition tasks on laparoscopic
  videos. IEEE transactions on medical imaging  \textbf{36}(1),  86--97 (2016)

\bibitem{vercauteren2019cai4cai}
Vercauteren, T., Unberath, M., Padoy, N., Navab, N.: Cai4cai: the rise of
  contextual artificial intelligence in computer-assisted interventions.
  Proceedings of the IEEE  \textbf{108}(1),  198--214 (2019)

\bibitem{wang2018videos}
Wang, X., Gupta, A.: Videos as space-time region graphs. In: ECCV. pp. 399--417
  (2018)

\bibitem{wu2021cross}
Wu, J.Y., Tamhane, A., Kazanzides, P., Unberath, M.: Cross-modal
  self-supervised representation learning for gesture and skill recognition in
  robotic surgery. IJCARS  \textbf{16}(5),  779--787 (2021)

\bibitem{Zhang_2022_ACCV}
Zhang, C., Gupta, A., Zisserman, A.: Is an object-centric video representation
  beneficial for transfer? In: Proceedings of the Asian Conference on Computer
  Vision (ACCV). pp. 1976--1994 (December 2022)

\bibitem{zhang2019graph}
Zhang, S., Tong, H., Xu, J., Maciejewski, R.: Graph convolutional networks: a
  comprehensive review. Computational Social Networks  \textbf{6}(1),  1--23
  (2019)

\end{thebibliography}

\newpage

\title{Encoding Surgical Videos as Latent Spatiotemporal Graphs for Object and Anatomy-Driven Reasoning: Supplementary Material}
\titlerunning{Encoding Surgical Videos as Spatiotemporal Graphs: Supplementary}
\author{}
\institute{}
\maketitle

\begin{table}[]
\centering
\caption{CVS Clip Classification Performance for various clip lengths.}
\renewcommand{\arraystretch}{1.5}
\begin{tabular*}{\linewidth}{@{\extracolsep{\fill}} ccccc }
\multirow{2}{*}{Clip Size} & \multicolumn{3}{c}{Performance (CVS mAP)}                      \\
                           & DeepCVS-Temporal & STRG          & Ours          \\ \hline
$5$                          & $62.39 \pm 1.63$    & $60.07 \pm 1.46$ & $63.70 \pm 1.12$ \\
$10$                         & $63.75 \pm 2.12$    & $60.72 \pm 1.74$ & $\mathbf{63.98 \pm 0.91}$ \\
$15$                         & $63.64 \pm 2.50$    & $59.35 \pm 1.82$ & $63.44 \pm 1.09$
\end{tabular*}
\end{table}

\begin{table}[]
\centering
\caption{Performance of DeepCVS-Temporal with various temporal models.}
\renewcommand{\arraystretch}{1.5}
\renewcommand{\tabcolsep}{1cm}
\begin{tabular}{cc}
Temporal Model & Performance (CVS mAP) \\ \hline
LSTM & $63.13 \pm 3.17$ \\
GRU & $63.26 \pm 2.21$ \\
Transformer & $\mathbf{63.75 \pm 2.12}$
\end{tabular}
\end{table}

\begin{table}[]
\centering
\caption{Hyperparameter Settings for Our Model.}
\begin{tabular*}{\linewidth}{@{\extracolsep{\fill}} ccc }
\multirow{2}{*}{Setting} & \multicolumn{2}{c}{Value} \\
 & CVS Clip Classification & Video Phase Recognition \\ \hline
Clip Length & 10 & Variable (Video Length) \\
Clip FPS & 1 & 0.5 \\
Optimizer & Adam & Adam \\
Learning Rate & 3e-4 & 1e-3 \\
Batch Size & 128 & 1 \\
Dropout & 0.25 & 0.0 \\
Label Per-Frame & Yes & Yes
\end{tabular*}
\end{table}

\begin{figure}
    \centering
    \includegraphics[width=\textwidth]{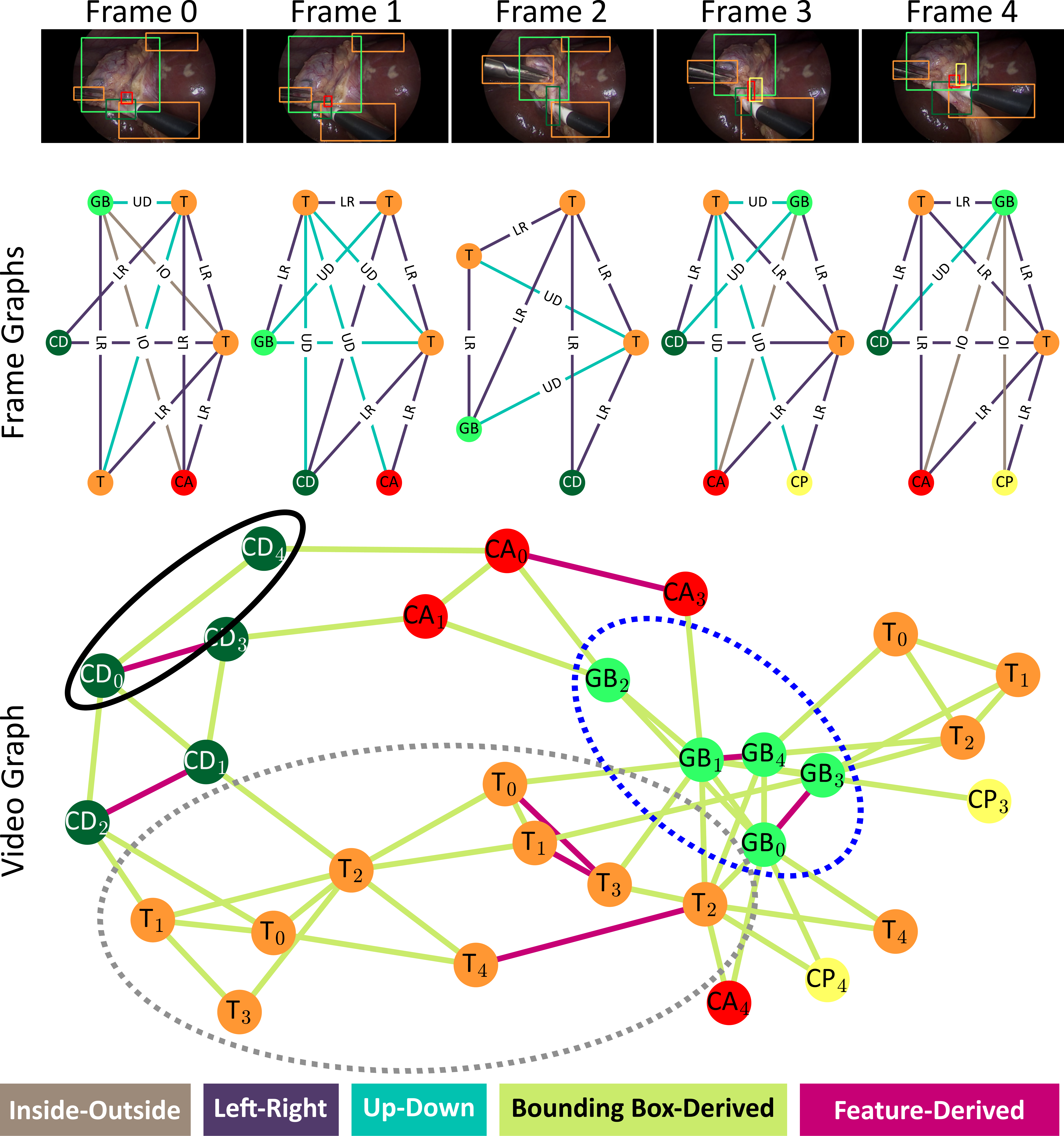}
    \caption{Predicted frame-wise and spatiotemporal graph structure for a 5-frame clip. Subscripts indicate the frame that each node comes from. The blue circled region shows a well tracked object forming a cluster in the video graph. The black circled region shows the impact of adding edges at multiple temporal horizons, as CD$_4$ would otherwise be unconnected to the other CD instances. Lastly, the gray circled region shows the importance of feature-derived edges; two sets of tool instances are part of the same cluster, but feature-derived edges help distinguish the different instances. Legend - CT: Calot Triangle, CP: Cystic Plate, CA: Cystic Artery, CD: Cystic Duct, GB: Gallbladder, T: Tool. Differently colored edges correspond to different relation classes.}
    \label{fig:qual}
\end{figure}

\end{document}